\documentclass[10pt]{article}
\pdfoutput=1

\usepackage{graphicx} 
\usepackage{subfigure} 

\usepackage{algorithm}
\usepackage{algorithmic}
\usepackage{amsmath,amsthm}
\usepackage{slashbox}
\usepackage{multirow}
\usepackage{amssymb}
\usepackage{hyperref}

\begin{document}


\title{{\bf Deep Transductive Semi-supervised Maximum Margin Clustering}}
\author{Gang Chen
}
\maketitle

\vskip 0.3in

\begin{abstract} 
Semi-supervised clustering is an very important topic in machine learning and computer vision. The key challenge of this problem is how to learn a metric, such that the instances sharing the same label are more likely close to each other on the embedded space. However, little attention has been paid to learn better representations when the data lie on non-linear manifold. Fortunately, deep learning has led to great success on feature learning recently. Inspired by the advances of deep learning, we propose a deep transductive semi-supervised maximum margin clustering approach. 
More specifically, given pairwise constraints, we exploit both labeled and unlabeled data to learn a non-linear mapping under maximum margin framework for clustering analysis. Thus, our model unifies transductive learning, feature learning and maximum margin techniques in the semi-supervised clustering framework. We pretrain the deep network structure with restricted Boltzmann machines (RBMs) layer by layer greedily, and optimize our objective function with gradient descent. By checking the most violated constraints, our approach updates the model parameters through error backpropagation, in which deep features are learned automatically. The experimental results shows that our model is significantly better than the state of the art on semi-supervised clustering.  
\end{abstract} 

\section{Introduction}\label{intro}
In this paper, we investigate the semi-supervised clustering with side information in the form of pairwise constraints. In general, a pairwise constraint between two examples indicates whether they belong to the same cluster or not, which provides the supervision information: a same-label (or must-link) constraint denotes that the pair of instances should be partitioned into the same cluster, while a different-label (or cannot-link) constraint specifies that the pair of instances should be assigned into different clusters \cite{Xing03,Shalev-Shwartz04,Weinberger09}.

Semi-supervised learning with pairwise constraints, has received considerable attention recently, especially for classification and clustering \cite{Xing03,Bilenko04,Goldberger04,Shalev-Shwartz04,Davis07,Weinberger09,Zeng12}. On the one hand, it is relatively easy to decide whether two items are similar or not from human in the loop because it often involves little effort from users. 
On the other hand, the maximum margin techniques have shown promising performance on classification tasks, and thus it has been widely used in semi-supervised clustering \cite{Vapnik1995,Tsochantaridis05,Zhang07,Weinberger09,Zeng12}. In general, traditional semi-supervised clustering approaches either learn a distance metric based on the pairwise constraints, or leverage discriminative methods, such as k-nearest neighbor (kNN) and support vector machines (SVM) for better clustering performance. 
However, to collapse examples that belong to the same cluster approximately into a single point cannot always be achieved with simple linear transformations, especially when the data lie on an non-linear manifold. Although kernel methods are widely used for non-linear cases, it is a shallow approach and needs to specify hyper parameters in most situations \cite{Vapnik1995,Weinberger09}. Fortunately, recent advances in the training of deep networks provide a way to learn non-linear transformations of data, which are useful for supervised/unsupervised tasks \cite{Erhan10,Bengio12}.

Inspired  by feature learning \cite{Hinton06a,Vincent10,Bengio12}, we propose a deep transductive semi-supervised clustering approach, which inherits both advantages from deep learning and maximum margin methods. Our method can learn features automatically from observation, kind of learning a metric as in \cite{Weston08}. However, unlike the linear mapping, e.g. Mahalanobis metric \cite{Weinberger09,Zeng12}, our method can learn a non-linear manifold representation, which is helpful for clustering and classification \cite{Bengio12}. With the learned features as the input to the semi-supervised maximum margin clustering framework, we can learn the clustering weights. To leverage the unlabeled data, we also incorporate transductive learning to improve the clustering analysis. Through backpropagation, our approach can learn discriminative features via maximum margin techniques. Hence, our model unifies maximum margin, semi-supervised information and deep learning in an joint framework. We pre-train our model with stacked RBMs for feature representations firstly. And then we compute the gradient w.r.t. parameters and optimize our objective function in an alternative manner: data representation and model weights optimization with gradient descent. We test our model over a bunch of data sets and show that it yields accuracy significantly better than the state of the art.

The outline of this paper is as follows. In Section 2, we review the related work. Then, we present the model in Section 3. Section 4 present results of our experiments with the new techniques on a few widely used data sets. Finally we conclude the paper.
\section{Related work}
The semi-supervised clustering with partial labels generally explores two directions to improve performance: (1) leverage more sophisticated classification models, such as maximum margin techniques \cite{Tsochantaridis05,Weinberger09};
(2) learn a better distance metric \cite{Shalev-Shwartz04,Weinberger09}.

The maximum margin clustering (MMC) aims to find the hyperplanes that can partition the data into different clusters over all possible labels with large margins \cite{Xu05,Valizadegan06,Zhang07}. Nevertheless, the accuracy of the clustering results by MMC may not be good sometimes due to the nature of its unsupervised learning \cite{Hu08}. Thus, it is interested to incorporate semi-supervised information, e.g. the pairwise constraints, into the recently proposed maximum margin clustering framework. 
Recent research demonstrates the advantages by leveraging pairwise constraints on the semi-supervised clustering problems \cite{Wagstaff01,Klein02,Xing03,Bar-Hillel03,Cohn03,Bilenko04}. In particular, COPKmeans [11] is a semi-supervised variant of Kmeans, by following the same clustering procedure of Kmeans while avoiding violations of pairwise constraints. MPCKmeans \cite{Bilenko04} extended Kmeans and utilized both metric learning and pairwise constraints in the clustering process. More recently, \cite{Nguyen08} show that they can improve classification with pairwise constraints under maximum margin framework. \cite{Zeng12} leverage the margin-based approach on the semi-supervised clustering problems, and yield competitive results. 

How to learn a good metric over input space is critical for a successful semi-supervised clustering approach. Hence, another direction for clustering is to learn a distance metric \cite{Xing03,Shalev-Shwartz04,Goldberger04,Globerson05,Davis07,Weinberger09} which can reflect the underlying relationships between the input instance pairs. 
The pseudo-metric \cite{Shalev-Shwartz04} parameterized by positive semi-definite matrices (PSD) is learned with an online updating rule, that alternates between projections onto PSD and onto half-space constraints imposed by the instance pairs. \cite{Xing03} proposed to learn a distance metric (Mahalanobis) that respects pairwise constraints for clustering. In \cite{Davis07}, an information-theoretic approach to learning a Mahalanobis distance function via LogDet divergence is proposed. Recently, a supervised approach to learn Mahalanobis metric is also proposed in \cite{Weinberger09}, by minimizing the pairwise distances between instances in the same cluster, while increasing the separation between data points with dissimilar classes. To handle the data that lies on non-linear manifolds, kernel methods are widely used. Unfortunately, these non-linear embedding algorithms for use is shallow methods. 

On the other hand, recent advances in deep learning \cite{Hinton06a,Vincent10,Bengio12} have sparked great interest in dimension reduction \cite{Hinton06b,Weston08} and classification problems \cite{Hinton06a,Larochelle12}. In a sense, the success of deep learning lies on learned features, which are useful for supervised/unsupervised tasks \cite{Erhan10,Bengio12}. For example, the binary hidden units in the discriminative Restricted Boltzmann Machines (RBMs) \cite{Larochelle08,Gelfand10} can model latent features of the data that improve classification. 
The deep learning for semi-supervised embedding \cite{Weston08} extends shallow semi-supervised learning techniques such as kernel methods with deep neural networks, and yield promising results. The work of \cite{Tang13} is most related to our proposed algorithm. It presented deep learning with support vector machines, which can learn features under discriminative learning framework automatically with labeled data. However, their approach is totally supervised and for classification problems, while our model is for semi-supervised clustering problems. Compared to conventional methods, our model consider both feature learning and transductive principles in our semi-supervised clustering model, so that it can handles complex data distribution and learns a better non-linear mapping to improve clustering performance. 

\section{Deep Transductive Semi-supervised Maximum Margin Clustering}
In this section, we will introduce the transductive semi-supervised maximum margin clustering, with deep features learned simultaneously in an unified framework. 
\subsection{Overview of our approach}
Let $\mathcal{X} = \{ {\bf x}_{i} \}_{i=1}^N$ (${\bf x}_{i}  \in \mathbb{R}^D$) be a set of $N$ examples, which belongs to $K$ clusters called $\mathcal{Z}$. In addition to the unlabeled data, there is additional partially labeled data in the form of pairwise constraints $C =\{({\bf x}_i, {\bf x}_j, \delta(z_i = z_j)\}$, which is a kind of side information to provide whether the two instances $({\bf x}_i, {\bf x}_j)$ are from the same cluster or not (indicated by the $\delta$ function). Most methods attempt to learn weights ${\bf w}^k \in \mathbb{R}^D$, for each cluster $k = [1, K]$, to make these constraints satisfied as much as possible.

Instead of learning a linear mapping or Mahalanobis metric \cite{Shalev-Shwartz04,Weinberger09}, we are interested in a non-linear mapping function. To make it easy to understand, suppose we have learned a nonlinear mapping function $f:  \mathbb{R}^D \rightarrow \mathbb{R}^d $. Then, for each instance ${\bf x}  \in \mathcal{X}$, we can get its embedding code ${\bf h} = f({\bf x})$ (note that the pairwise constrains also are kept in the coding space). Then given the learned features ${\bf h} $, we leverage semi-supervised maximum margin clustering to partition the data. Just like the multi-class classification problems \cite{Tsochantaridis05}, we use the joint feature representation $\Phi({\bf h}, z)$ for each $({\bf h},z) \in \mathcal{X}\times \mathcal{Z}$ 

\begin{equation}
\centering
\Phi({\bf h}, z) = \begin{bmatrix}
      & {\bf h} \cdot \delta(z =1)        \\[0.3em]
      &      \cdot \cdot \cdot    \\[0.3em]
      & {\bf h} \cdot \delta(z =K) 
     \end{bmatrix}
\label{eq:feats}
\end{equation}
where $\delta$ is the indicator function (1 if the equation holds, otherwise 0). Correspondently, the hyperplanes for the $K$ clusters can be parameterized by the weight vector ${\bf W} \in \mathbb{R}^{(K\times d) \times 1}$, which is the concatenation of weights ${\bf w}^k$, for $k =\{1,...,K\}$. In other words, ${\bf W}[(k -1)\times d + 1:  k \times d ] = {\bf w}^k $. The clustering of testing examples is done in the same manner as the multiclass SVM \cite{Tsochantaridis05},
\begin{equation}
\max_{z\in [1,K]} {\bf W}^T \Phi (  f({\bf x}), z) 
\label{eq:inference}
\end{equation}
For inference, we first project data into hidden space with function $f$ and then do clustering analysis. The problem left is how to learn the weight parameter ${\bf W}$ and the projection function $f$. 

\subsection{Objective function} \label{sec:obj}
We would like to extend semi-supervised clustering with deep feature learning. Deep learning consists of learning a model with several layers of non-linear mapping. 
As mentioned before, ${\bf h} \in \mathbb{R}^d$ is the mapping code with function $f$, which is non-linear mappings defined with $L$-layers neural network, s.t. 
\begin{equation}\label{eq:hidden}
{\bf h}_{i} =f({\bf x}_i) =  \underbrace{f_{L} \circ f_{L-1}\circ \cdot\cdot\cdot \circ f_{1}}_{L \textrm{ times}}({\bf x}_{i})
\end{equation}
where $\circ$ indicates the function composition, and $f_l$ is logistic function with the weight parameter $\boldsymbol{\theta}_l$ respectively for each layer $l = \{1,..,L\}$, refer further to Sec. \ref{sec:learning} for more details. With a little abuse of symbols, for any input ${\bf x}$, If we denote the output of the $l$-th layer as $f_{1\rightarrow l}( {\bf x})$, then we can get ${\bf h} = f_{1\rightarrow L}( {\bf x})$.

In a similar manner as in \cite{Nguyen08,Zeng12}, we will incorporate the pairwise constraint information into the margin-based clustering framework. In addition, we leverage the unlabeled data to separate clusters in large margins, by following transductive learning. Specifically, given the pairwise constraint set $C = \{ ({\bf x}_i , {\bf x}_j , \delta(z_i=z_j)) \}$, we first project the dataset $\mathcal{X}$ into embedded space and minimize the following transductive semi-supervised objective function
\begin{align} \label{eq:obj}
& \min_{{\bf W}, \boldsymbol{\Theta}} \frac{\lambda}{2} || {\bf W}||^2 +  \frac{1}{n^+} \sum_i \eta_{i}^+  + \frac{1}{n^-} \sum_j \eta_{j}^- +\frac{\beta}{UK} \sum_{i\in U} \xi_{i} \\
& s.t.   \nonumber  \\
& \forall s_{i1}, s_{i2} \in \mathcal{Z}, s_{i1} \neq s_{i2}; \textrm{if  }({\bf h}_{i1}, {\bf h}_{i2}, \delta(z_{i1}, z_{i2})) \in C^+ \nonumber \\
 & \max_{z_{i1} =z_{i2}} {\bf W}^T\Phi({\bf h}_{i1}, {\bf h}_{i2},z_{i1}, z_{i2} ) -  \nonumber \\
 & \quad \quad \quad {\bf W}^T\Phi({\bf h}_{i1}, {\bf h}_{i2}, s_{i1}, s_{i2}) \geq 1- \eta_i ,   \eta_i   \geq 0   \label{eq:cond1} \\
& \forall s_{j1}, s_{j2} \in \mathcal{Z}, s_{j1} = s_{j2}; \textrm{if  }({\bf h}_{j1}, {\bf h}_{j2}, \delta(z_{j1}, z_{j2})) \in C^- \nonumber \\
& \max_{z_{j1} \neq z_{j2}} {\bf W}^T\Phi({\bf h}_{j1}, {\bf h}_{j2}, z_{j1}, z_{j2}) -  \nonumber \\
&  \quad \quad \quad {\bf W}^T\Phi({\bf h}_{j1}, {\bf h}_{j2}, s_{j1}, s_{j2}) \geq 1- \eta_j  ,  \eta_j \geq 0   \label{eq:cond2} \\
& \forall i \in U, \forall s_i \neq z_{i}   \in \mathcal{Z}                                   \nonumber \\
&  \max_{z_{i} } {\bf W}^T\Phi({\bf h}_{i}, z_{i}) -   {\bf W}^T\Phi({\bf h}_{i}, s_{i}) \geq 1 - \xi_{i} \label{eq:cond3} 
\end{align}
where ${\bf W}$ is the clustering weight in the over the learned feature space, $\boldsymbol{\Theta}=  \{  \boldsymbol{\theta}_l\}_{l=1}^L$ are the weights for each layer in the deep architecture, and ${\bf h}_i$ is the mapping code from ${\bf x}_i$ via Eq. \ref{eq:hidden}; $C^+ = \{ ( {\bf h}_i, {\bf h}_j, \delta(z_i = z_j) ) | z_i=z_j\}$ are the same label pairs, with the total number of pairwise constraints $n^+ = |C^+|$, 
$C^- = \{ ( {\bf h}_i, {\bf h}_j, \delta(z_i = z_j) ) | z_i \neq z_j \}$ are different-label pairs, with $n^- = |C^-|$. $U$ is the number of the unlabeled data (instances), not belong to any pairwise constrains. For convenience, we define $ \Phi ({\bf h}_i, {\bf h}_j, z_i, z_j) = \Phi({\bf h}_i, z_i) + \Phi({\bf h}_j, z_j)$, which means the mapping of a pairwise constraint as the sum of the individual example-label mappings. The multi-layers non-linear mapping function $f$ projects ${\bf x}_i$ into ${\bf h}_i$, for $i\in[1, N]$. Instead of a linear mapping, the advantage of using a deep network to parametrize the function $f$ is that a multi-layer network is better at learning a non-linear function that is presumably required to collapse classes in the latent space, in particular when the data consists of very complex non-linear structures.

Eqs. \ref{eq:cond1} and \ref{eq:cond2} specify the conditions that need to be satisfied, which means that the score for the most possible assigning scheme satisfying the constraints should be greater than that for any other assigning scheme with large margins. More specifically, for any pair $({\bf h}_i, {\bf h}_j, 1)  \in C^+$, it requires that the largest score for assigning $({\bf h}_i, {\bf h}_j)$ into the same cluster should be greater than that for assigning the pair into different clusters by at least 1 (soft margin can be applied here too). Analogously, for any dissimilar pair $({\bf h}_i, {\bf h}_j, 0) \in C^-$, the score that they are assigned into the most two different clusters should be greater than that for partitioning them into the same cluster. 

Eq. \ref{eq:cond3} is from the principles of transductive learning, which indicates that the score of the most assigned cluster label is greater at least 1 than that of the runner up from the rest clusters.


The constrained optimization problem in Eq. \ref{eq:obj} is hard to solve because the first inequality Eq. \ref{eq:cond1} and the second inequality Eq. \ref{eq:cond2} impose all the possible combinations of two clusters for each pairwise constraint. 
Thus, we transform it into the following equivalent unconstrained function which it is generally easier to solve 
\begin{subequations}\label{eq:eqe2}
\begin{align}
& \min \frac{\lambda}{2} || {\bf W}||^2 \nonumber \\
 &+ \frac{1}{n^+}  \bigg \{1- \bigg[\max_{\substack{z_{i1} =z_{i1} \\ ({\bf h}_{i1}, {\bf h}_{i2},1) \in C^+}} {\bf W}^T\Phi({\bf h}_{i1}, {\bf h}_{i2},z_{i1}, z_{i2} ) -  \nonumber \\
 & \quad \quad \quad \max_{s_{i1} \neq s_{i1}} {\bf W}^T\Phi({\bf h}_{i1}, {\bf h}_{i2}, s_{i1}, s_{i2}) \bigg ]  \bigg \}_{+}  \label{eq:cond4} \\
&+ \frac{1}{n^-}  \bigg \{1-   \bigg[  \max_{\substack{z_{j1} \neq z_{j2}  \\  ({\bf h}_{j1}, {\bf h}_{j2},0) \in C^- }} {\bf W}^T\Phi({\bf h}_{j1}, {\bf h}_{j2}, z_{j1}, z_{j2}) -  \nonumber \\
&  \quad \quad \quad \max_{s_{j1} = s_{j2}} {\bf W}^T\Phi({\bf h}_{j1}, {\bf h}_{j2}, s_{j1}, s_{j2}) \bigg ]  \bigg \}_{+}      \label{eq:cond5}   \\
& + \frac{\beta}{UK} \sum_{i\in U}  \big\{ 1 - [\max_{z_{i} } {\bf W}^T\Phi({\bf h}_{i}, z_{i}) -   \max_{s_{i} \neq z_{i}} {\bf W}^T\Phi({\bf h}_{i}, s_{i}) ] \big\}_{+}
\end{align}
\end{subequations}
where $\{ x\}_+  = max(x, 0)$ and ${\bf h}_i$ is the projected code of ${\bf x}_i$ using Eq. \ref{eq:hidden}. The formula \ref{eq:cond4} specifies the condition that need to be satisfied for the same label pairwise constrains, while formula \ref{eq:cond5} denotes the conditions for different-label pairs. The last equation is corresponding to transductive constraints in Eq. \ref{eq:obj}. 

In the objective function, we need to estimate the parameters, the weight ${\bf W}$, as well as the weights ${\boldsymbol{\theta}}_l$ for each layer $l \in [1, L]$ in the deep network. From the objective function, we can compute the gradients w.r.t. ${\bf W}$ and ${\boldsymbol{\theta}}_l$ for $l \in [1, L]$ (via backpropagation) respectively, and gradient-based methods can be used to optimize it. Note that ${\bf h}$ (we ignore the subscript for convenience) in the objective function Eq. \ref{eq:eqe2} is the non-linear embedding code from ${\bf x}$. Thus, the objective function is not convex anymore, and we can only find a local minimum. In practice, we find L-BFGS cannot work well and easily trap into a bad local minimum. In our work, we use (sub)gradient descent to optimize the objective function, by projecting the training data with $f$ and optimizing the objective function in an alternative manner. 
\subsection{Parameter learning} \label{sec:learning}
We learn the parameters in an alternative manner: (1) data projection, given the model parameters; (2) and then optimize model parameters with gradient descent. 
To compute the gradients of the parameters, we need to find the most violated constraints first. For the same label pairs, we have the following most violated set:
\begin{align}\label{eq:setpos}
 A^+=   \bigg \{ ({\bf h}_i, {\bf h}_j, \delta(z_{i1} = z_{i2})) \in C^+ | 
  \max_{\substack{z_{i1} =z_{i1} }} {\bf W}^T\Phi({\bf h}_{i1}, {\bf h}_{i2},z_{i1}, z_{i2} )   
- \max_{s_{i1} \neq s_{i1}} {\bf W}^T\Phi({\bf h}_{i1}, {\bf h}_{i2}, s_{i1}, s_{i2}) <1  \bigg \} 
\end{align}
For the different-label pairs, we denote the most violated set as
\begin{align}
A^-   = \bigg \{  ({\bf h}_i, {\bf h}_j, \delta(z_{j1} = z_{j2})) \in C^- |  
\max_{\substack{z_{j1} \neq z_{j2}  }} {\bf W}^T\Phi({\bf h}_{j1}, {\bf h}_{j2}, z_{j1}, z_{j2}) 
- \max_{s_{j1} = s_{j2}} {\bf W}^T\Phi({\bf h}_{j1}, {\bf h}_{j2}, s_{j1}, s_{j2}) < 1  \bigg \}
\label{eq:setneg}
\end{align}
Then, we compute the gradient w.r.t. ${\bf W}$
\begin{align} \label{eq:gradient}
& d{\bf W} = \lambda {\bf W} + \nonumber \\
&-  \frac{1}{n^+}    \sum_{({\bf h}_{i1}, {\bf h}_{i2}, 1) \in A^+} \bigg[ \Phi({\bf h}_{i1}, {\bf h}_{i2},z_{i1}^+, z_{i2}^+ ) - \Phi({\bf h}_{i1}, {\bf h}_{i2}, z_{i1}^-, z_{i2}^-) \bigg]  \nonumber \\
&-    \frac{1}{n^-}    \sum_{({\bf h}_{j1}, {\bf h}_{j2}, 0) \in A^-}  \bigg[\Phi({\bf h}_{j1}, {\bf h}_{j2},z_{j1}^-, z_{j2}^- ) - \Phi({\bf h}_{j1}, {\bf h}_{j2}, z_{j1}^+, z_{j2}^+) \bigg] \nonumber  \\
& -  \sum_{i \in U}  \frac{\beta}{UK}\bigg[     \Phi({\bf h}_{i}, z_{i}^+) -   \Phi({\bf h}_{i}, s_{i}^+)   \bigg], 
\end{align} 
where $(z_{i1}^+, z_{i2}^+)  = \max_{\substack{z_{i1} =z_{i2} }} {\bf W}^T\Phi({\bf h}_{i1}, {\bf h}_{i2},z_{i1}, z_{i2} ) $,  
$(z_{i1}^-, z_{i2}^-)  = \max_{\substack{z_{i1} \neq z_{i1} }} {\bf W}^T\Phi({\bf h}_{i1}, {\bf h}_{i2},z_{i1}, z_{i2} )  $; and for the unlabeled set $z_{i}^+ = \max_{z_{i}} \Phi({\bf h}_{i}, z_{i}) $ and $s_{i}^+ = \max_{s_{i} \neq z_{i}^+} \Phi({\bf h}_{i}, s_{i}) $

In order to learn discriminative features, we also need to estimate the weights in the multi-layer network. Note that for each pair $({\bf h}_i, {\bf h}_j)$, if it violates the constraints in Eqs. \ref{eq:cond1} and \ref{eq:cond2}, then we can compute the gradient w.r.t. ${\bf h}_i$ and ${\bf h}_j$ respectively, which will be used to calculate the gradients of ${\boldsymbol{\theta}}_l$ for $l \in [1,L]$ in the deep network. We use ${\bf H} = [{\bf h}_1, {\bf h}_2, ..., {\bf h}_N]$ as the concatenation of all the hidden codes, where ${\bf H} \in \mathcal{R}^{d \times N}$, with each column ${\bf H}(:, i) = {\bf h}_i$.

For the positive pairs, we have 
\begin{subequations}\label{eq:pairpos}
\begin{align}
d {\bf h}_{i1} = - \frac{1}{n^+} \sum_{i1} [{\bf W}_{z_{i1}^+}    -   {\bf W}_{z_{i1}^-} ]   \label{eq:pairpos1}\\
d {\bf h}_{i2} =-  \frac{1}{n^+}  \sum_{i2} [ {\bf W}_{z_{i1}^+}    -   {\bf W}_{z_{i2}^-}  ]    \label{eq:pairpos2}
\end{align}
\end{subequations}
where ${\bf W}_{z_{i1}^+}$ indicates the weight vector corresponding to the cluster label $z_{i1}^+$ in the whole weight matrix ${\bf W}$. More specifically, ${\bf W}_{z_{i1}^+} = {\bf W}[(z_{i1}^+ -1)\times d + 1:  z_{i1}^+ \times d ]$

For the negative pairs, we can get
\begin{subequations}\label{eq:pairneg}
\begin{align}
d {\bf h}_{j1} = - \frac{1}{n^-} \sum_{j1} [{\bf W}_{z_{j1}^-}    -   {\bf W}_{z_{j1}^+} ]   \label{eq:pairneg1}\\
d {\bf h}_{j2} =-  \frac{1}{n^-}  \sum_{j2} [ {\bf W}_{z_{j2}^-}    -   {\bf W}_{z_{j1}^+}  ]  \label{eq:pairneg2}
\end{align} 
\end{subequations}

For the unlabeled instances, we have
\begin{equation}
d {\bf h}_{i} = - \frac{\beta}{UK} \sum_i  [ {\bf W}_{z_{i}^+}    -   {\bf W}_{s_{i}^+}  ]  
\label{eq:unlabeled}
\end{equation}
Given the gradient of $d{\bf h}_i$ for each hidden code, we can get the gradient w.r.t. ${\bf H}$ as 
\begin{equation}\label{eq:feats}
d {\bf H}(:, i) = d {\bf h}_{i}
\end{equation}
where $d{\bf h}_i$ can be calculated according to Eqs. \ref{eq:pairpos} and \ref{eq:pairneg}.
Then, we can calculate the gradients w.r.t. lower level weights with back-propagation. 
For example $d{\boldsymbol{\theta}}_{L} = d{\bf H} \times \big( f_{1\rightarrow L}( \mathcal{X} ) \cdot (1- f_{1\rightarrow L}( \mathcal{X} )) \big)$, where $\times$ representas matrix multiplication, and $\cdot$ indicates pointwise product.

\begin{algorithm}[t!]
\caption{} 
\label{alg1} 
\begin{algorithmic}[1]
\STATE Input: the training data $\mathcal{X}$, pairwise constraints ${C}$, the number of clusters $K$, the number of iterations $T$, $\lambda$, and $\beta$;
\STATE Initialize $\textbf{W}$;
\STATE Initialize ${\bf w}_l$ for $l = \{1,...,L\}$ layer-by-layer greedily;
\FOR{$i=1;  i<=T;  i++$}
\STATE if the objective in Eq. \ref{eq:eqe2} has no significant changes, break; 
\STATE project all training data $\mathcal{X}$ into latent space via Eq. \ref{eq:hidden};
\STATE find the most violated constrains according to Eqs. \ref{eq:setpos} and \ref{eq:setneg}
\STATE compute the gradient w.r.t. ${\bf W}$ via Eq. \ref{eq:gradient};
\STATE compute the gradient w.r.t. ${\bf H}$ via Eq. \ref{eq:feats};
\STATE compute the gradient w.r.t. ${\boldsymbol{\Theta}} = \{{\boldsymbol{\theta}}_l\}_{l=1}^L$ with backpropagation;
\STATE update the parameters with gradient descent via Eq. \ref{eq:descent};
\ENDFOR
\STATE Return model parameters $\textbf{W}$ and $\{{\bf w}_l\}_{l=1}^L$, as well as average accuracy;
\end{algorithmic}
\end{algorithm}

{\bf Initialization}: 
We used stacked RBMs to initialize the weights layer by layer greedily in the deep network, with contrastive divergence \cite{Hinton06a} (we used CD-1 in our experiments). Note that we used gaussian RBMs for the continuous data in the first layer, otherwise we used binary RBMs. 

In our deep model, the weights from the layers $1$ to $L$ are ${\boldsymbol{\theta}}_l$ respectively, for $l = \{1,..,L\}$, and the top layer $L$ has weight ${\boldsymbol{\theta}}_L$. We first pre-train the $L$-layer deep structure with RBMs layer by layer greedily. Thus, our deep network can learn parametric nonlinear mapping from input ${\bf x}$ to output ${\bf h}$, $f: {\bf x} \rightarrow {\bf h}$. 
Specifically, we think RBM is a 1-layer deep network, with weight ${\boldsymbol{\theta}}_1$. For example, for 1-layer DBN, we have ${\bf h} = f_1({\bf x})  = \textrm{logistic}({\boldsymbol{\theta}}_{1}^T [{\bf x}, 1])$, where we extend ${\bf x} \in \mathbb{R}^D$ into $[{\bf x},1] \in \mathbb{R}^{(D+1)}$ in order to handle bias in the non-linear mapping. Given the output of the current layer as the input to the next layer, we can learn each layer weight greedily. 

As for the clustering weight ${\bf W}$, we take a similar strategy as in \cite{Zeng12} to initialize it.

{\bf Parameter updating}: In our model, we use the gradient descent to update the model parameters. 
We also tried L-BFGS \cite{Byrd95,Rasmussen05} to update model parameters, but it did not perform well. 
In our model, we can update the model parameters as follows, 
\begin{align}
& {\bf W} \leftarrow  {\bf W}  - \gamma_{\bf W}    d{\bf W}, \nonumber \\
& {\boldsymbol{\theta}}_l \leftarrow  {\boldsymbol{\theta}}_l  - \gamma_{{\boldsymbol{\theta}}_l}    d{{\boldsymbol{\theta}}_l},  \text{$l \in \{1,...,L\}$}
\label{eq:descent}
\end{align}
where $\gamma_{\bf W}$ is the learning rate for the clustering weight ${\bf W}$, and $\gamma_{{\boldsymbol{\theta}}_l}$ is the learning rate for weights ${{\boldsymbol{\theta}}_l}$ in the deep neural network. Thus, our method alternates between data projection and parameter optimization. For more details, refer to algorithm \ref{alg1}.

After we learned the model parameters, we can do cluster analysis according to Eq. \ref{eq:inference}.

\section{Experiments} 
In this section, we presented a set of experiments comparing our method to the state of the art semi-supervised clustering methods on a wide range of data sets, including UCI data sets and the Reuters dataset in Table \ref{tab:dataset}, as well as the MNIST digits, COIL-20 and COIL-100 datasets. We also evaluated whether the transductive constraint in Eq. (\ref{eq:cond3}) is helpful or not in the clustering analysis.
\begin{table}
\centering
\begin{tabular}{lccc}
\hline
\multirow{2}{*}{Dataset} & \multicolumn{3}{c}{Description} \\
\cline{2-4}
    & \#vectors & dim & \#clusters \\
\hline
Glass        & 214        & 9    &   7\\
Wdbc       & 569     & 32    & 2\\
Wine       & 178     & 13     &  3 \\
Sonar      & 208    &  60 &   2 \\
Image Segmentation      & 2310    &  19  &   7 \\
Reuters & 8293    &  18933  &   65 \\
\hline
\end{tabular}
\caption{Descriptions of the UCI datasets and the Reuters dataset.}
\label{tab:dataset}
\end{table}

\subsection{Experimental setup}
In the experiments, we compared our method to the state of the art semi-supervised clustering approaches, including Xing \cite{Xing03}, ITML \cite{Davis07}, KISSME \cite{Koestinger12a} and CMMC \cite{Zeng12}. Note that Xing, ITML and KISSME are the semi-supervised approaches for metric learning (Mahalanobis). Thus, we used those methods to learn the metric and calculate the distances between all instances, then we used the kernel k-means \cite{Dhillon04} for clustering. Our method and CMMC are similar, which can be directly optimized for clustering. 

As for parameter setting, we set $\lambda = 0.02$ and $\beta = 1$. 
The learning rate $\gamma_{\bf W}$ decreases in the iterations in our model, by setting $\gamma_{\bf W}= \frac{1}{\lambda \times (i+1)}$, where $i$ is the index for iterations; while the learning rate for weighs in the deep network fixed, with $\gamma_{{\boldsymbol{\theta}}_l}=0.01$, for $l =\{1,...,L\}$. Without other specification, our model used the one hidden layer with 100 units on most data sets, except on the MNIST and UCI data sets. 

We tested our method on two tasks: pairwise classification and clustering analysis. As for pairwise classification, we randomly sampled 200 pairs of constraints (around 100 must-links and 100 cannot links), of which we used 100 pairwise constraints (50 must-links and 50 cannot links) as the training set, and the rest 100 pairs as the testing sets. Then we used the receiver operating characteristic (ROC) to evaluate the performance. 

As for the clustering analysis, we used the pairwise constraints sampled to train our model, then we use the learned model for clustering analysis. We used the accuracy (the most possible matching between the obtained labels and the original true labels, refer to \cite{Zeng12}) and adjusted Rand Index \cite{Hubert85,Rand71} to evaluate our method in all the experiments.

%
%
%

\begin{table*}[t!]
\centering
\resizebox{17.5cm}{!}{
\begin{tabular}{lcccccc|cccccc}
\hline
\multirow{2}{*}{Methods} & \multicolumn{6}{c|}{Accuracy (\%)} & \multicolumn{6}{c}{Adjusted Rand Index} \\
    & Glass & Wdbc & Wine & Sonar & Segmentation & Reuters & Glass & Wdbc & Wine & Sonar & Segmentation & Reuters\\
\hline
Xing \cite{Xing03}        & 46.2    & 91.9    & 81.5    &53.4    & 28.5     & 44.3          & 0.214    & 0.70    & 0.584    &0.02  & 0.12 & 0.14\\
ITML \cite{Davis07}      & 47.4     & {\bf 92.1}      & 70.2     & 69.2    & 30.0     & 45.0     & {\bf 0.223}     & {\bf 0.71}      & 0.520     & 0.14 & 0.14 & 0.15 \\
KISSME \cite{Koestinger12a}  & 36.5  & 77.9 & 65.2 &  67.3 &27.6    & 49.7         & 0.07       & 0.287     & 0.466      &  0.12 &0.09 & 0.17\\
CMMC \cite{Zeng12}      & 43.2     & 89.5     &  97.1      & 72.1  & 51.4 & 66.5       & 0.217     & 0.620     &  0.918     & 0.191  & 0.35 & 0.22\\
Our method                     & {\bf 50.9}  &  91.5   &   {\bf 98.8}      & {\bf 72.6}  & {\bf 57.1}     & {\bf 72.7}       & 0.219  &  0.689   &   {\bf 0.965}      & {\bf 0.20}  & {\bf 0.41} & {\bf 0.56}\\
\hline
\end{tabular}}
\caption{The experimental comparison on the UCI data sets and the Reuters data set. For the real UCI datasets, our method outperforms other methods significantly, except on the Glass and Wdbc data sets. Our method is remarkably better on the Reuters dataset, with both accuracy and Rand index. }
\label{tab:tab1}
\end{table*}

\subsection{Results}
\begin{figure}[htp]
  \centering
  \begin{tabular}{cc}
    \includegraphics[trim = 35mm 83mm 35mm 84mm, clip, width=6.5cm]{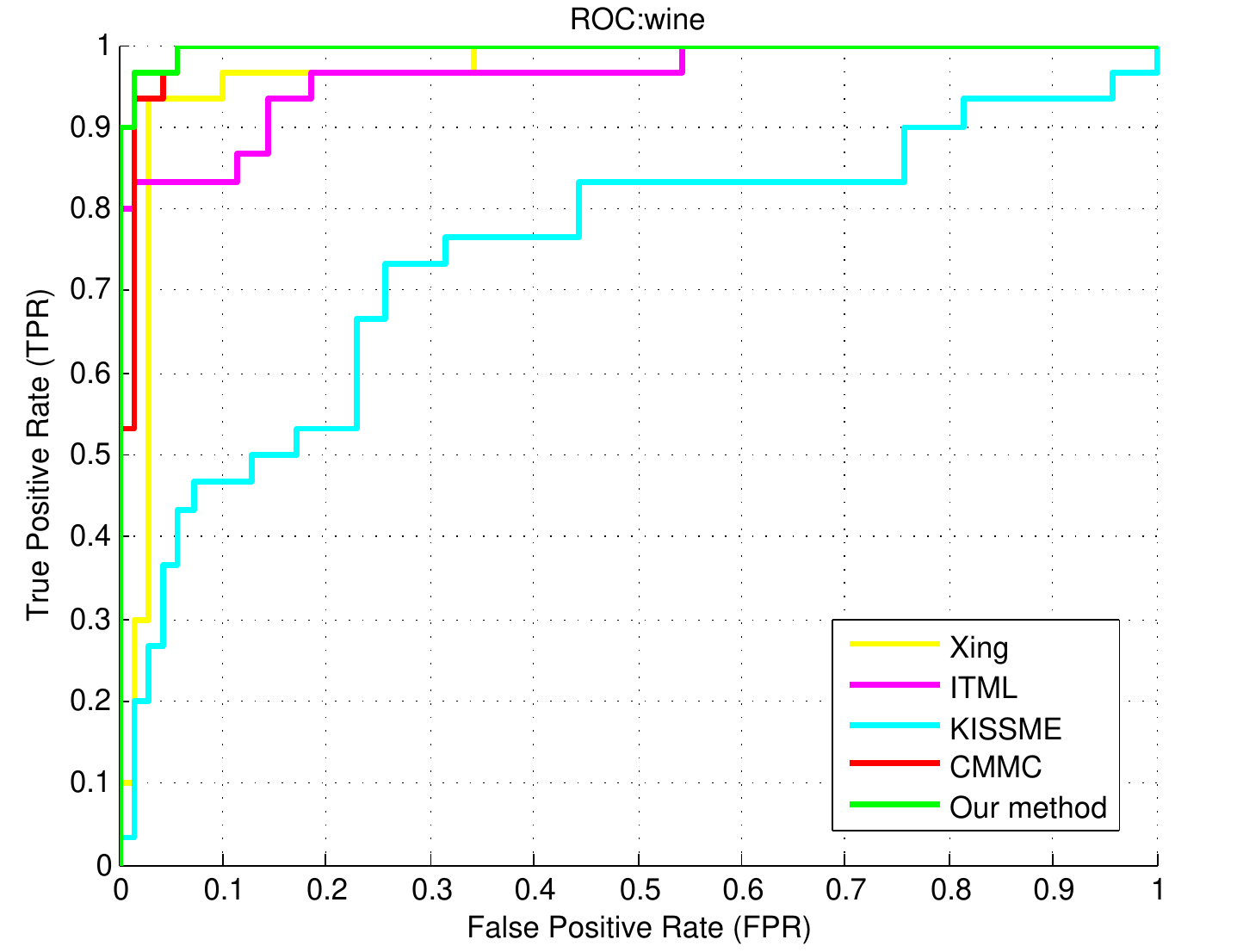}&
    \includegraphics[trim = 35mm 83mm 35mm 84mm, clip, width=6.5cm]{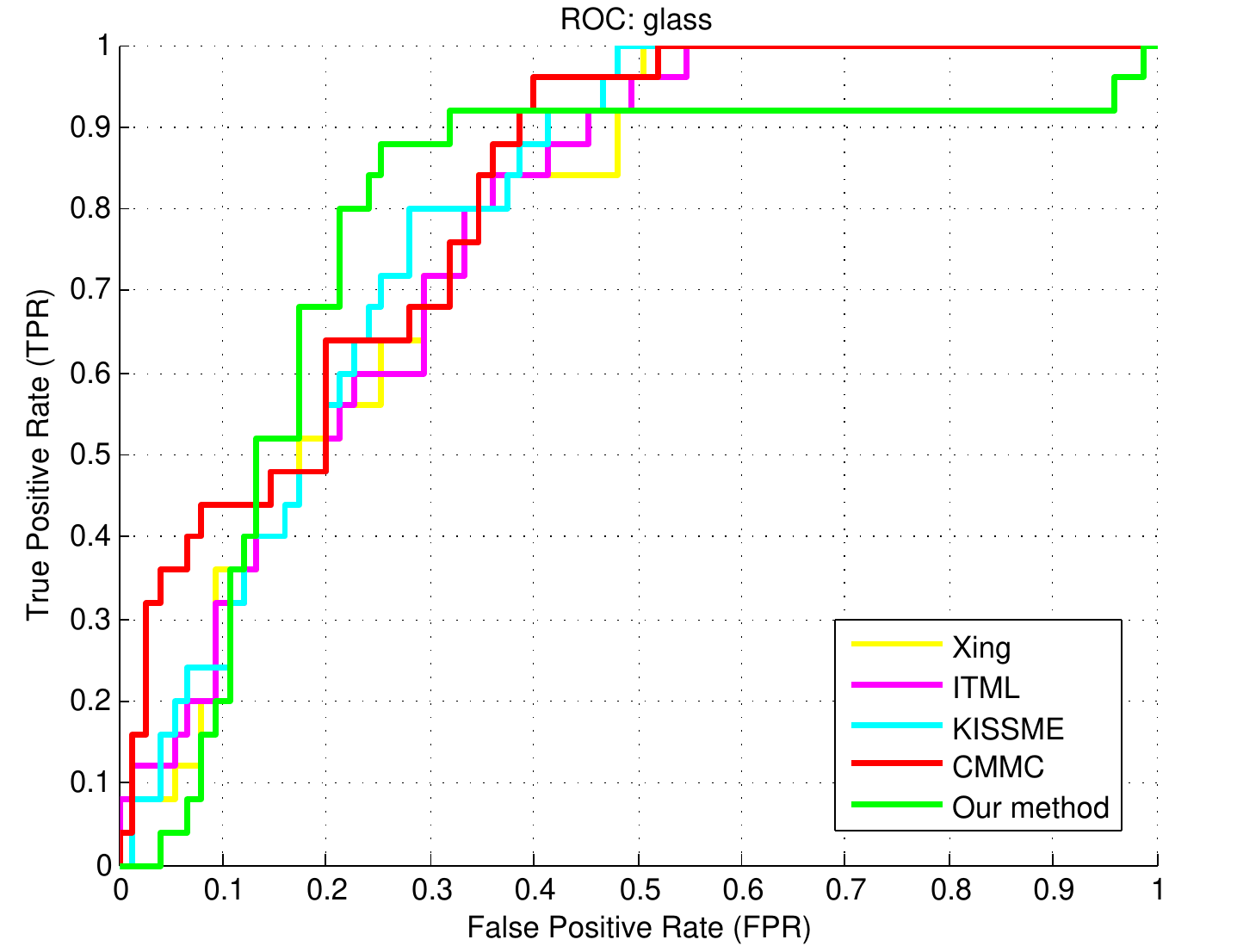}\\
    \includegraphics[trim = 35mm 83mm 35mm 84mm, clip, width=6.5cm]{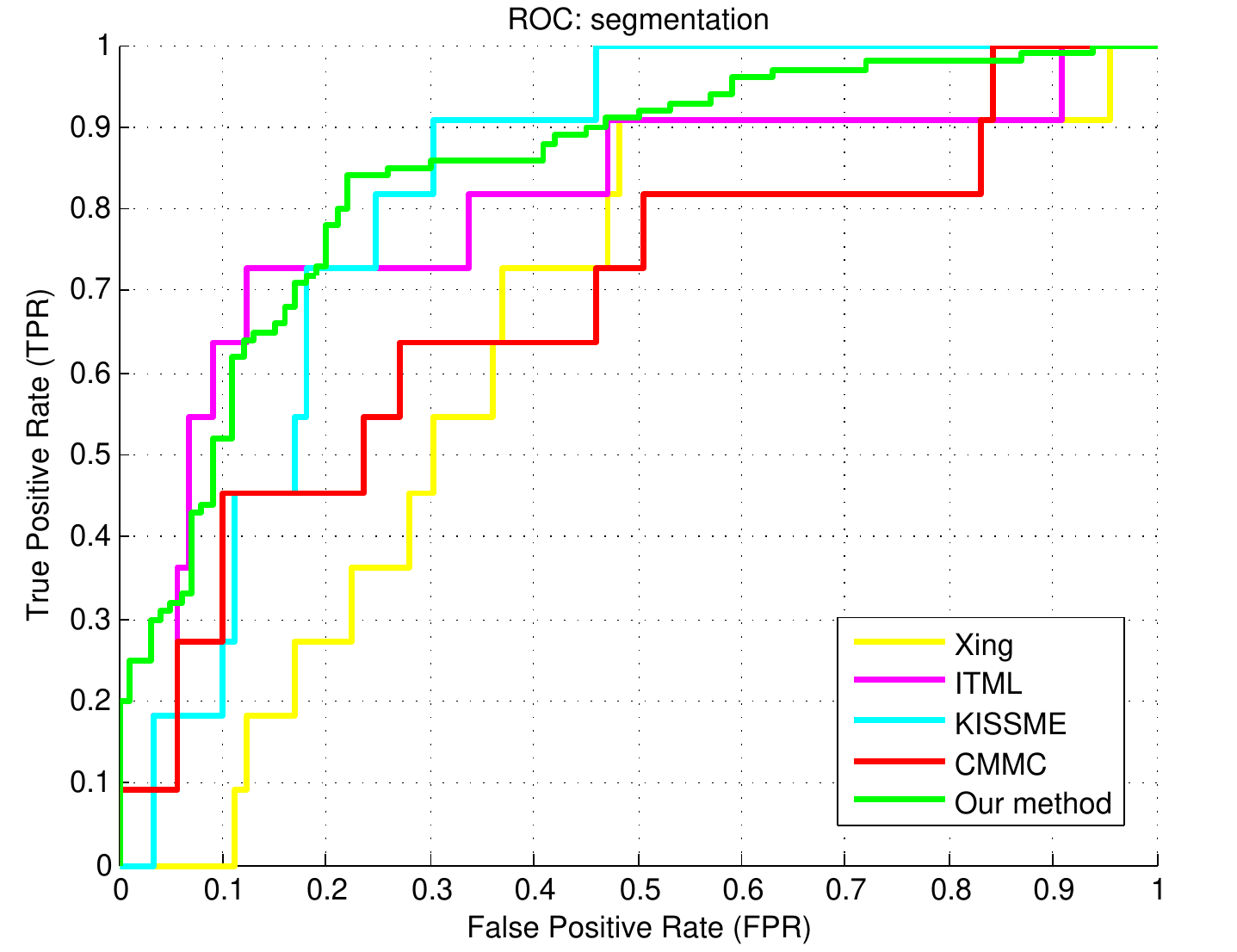}&
    \includegraphics[trim = 35mm 83mm 35mm 84mm, clip, width=6.5cm]{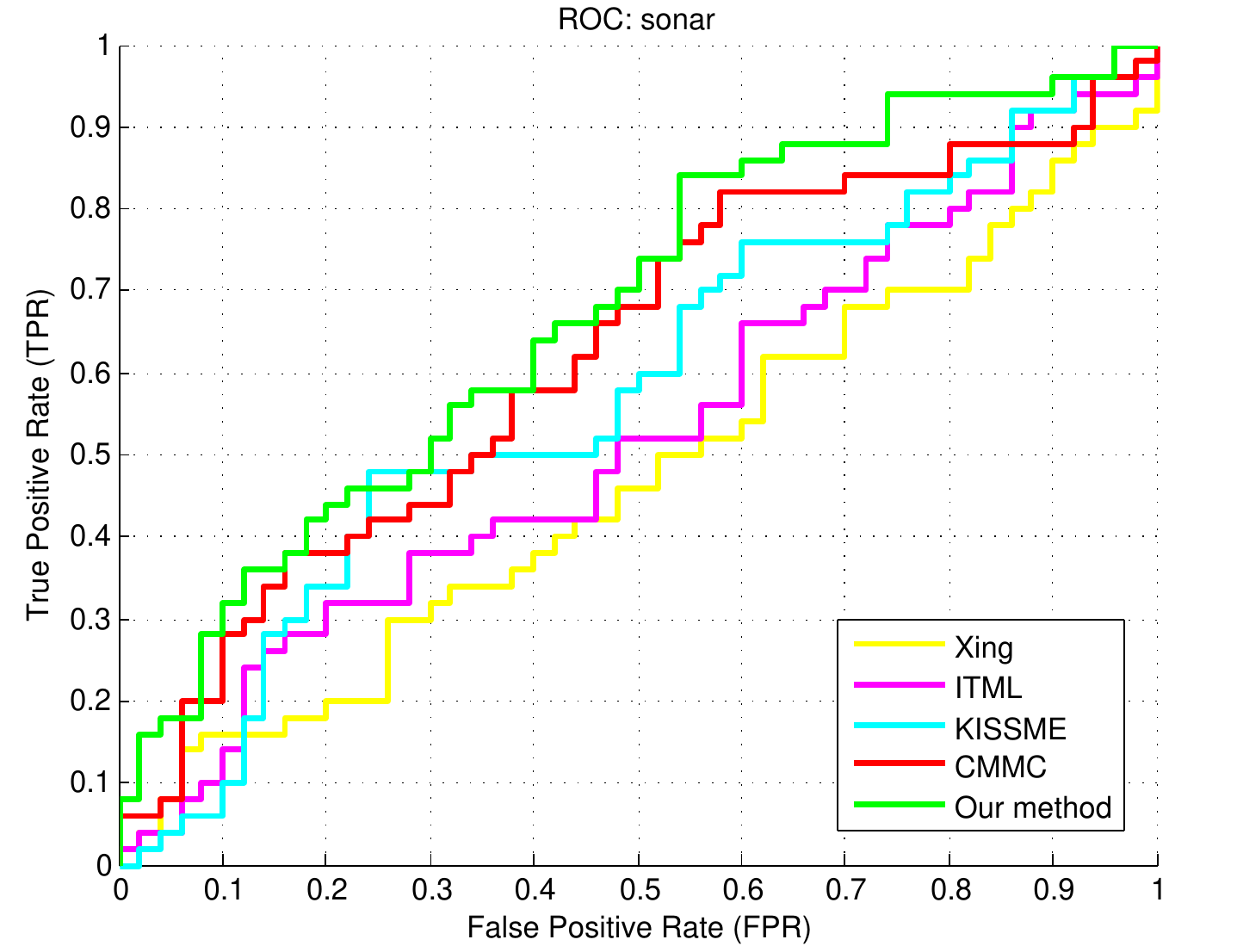}\\
    \includegraphics[trim = 35mm 83mm 35mm 84mm, clip, width=6.5cm]{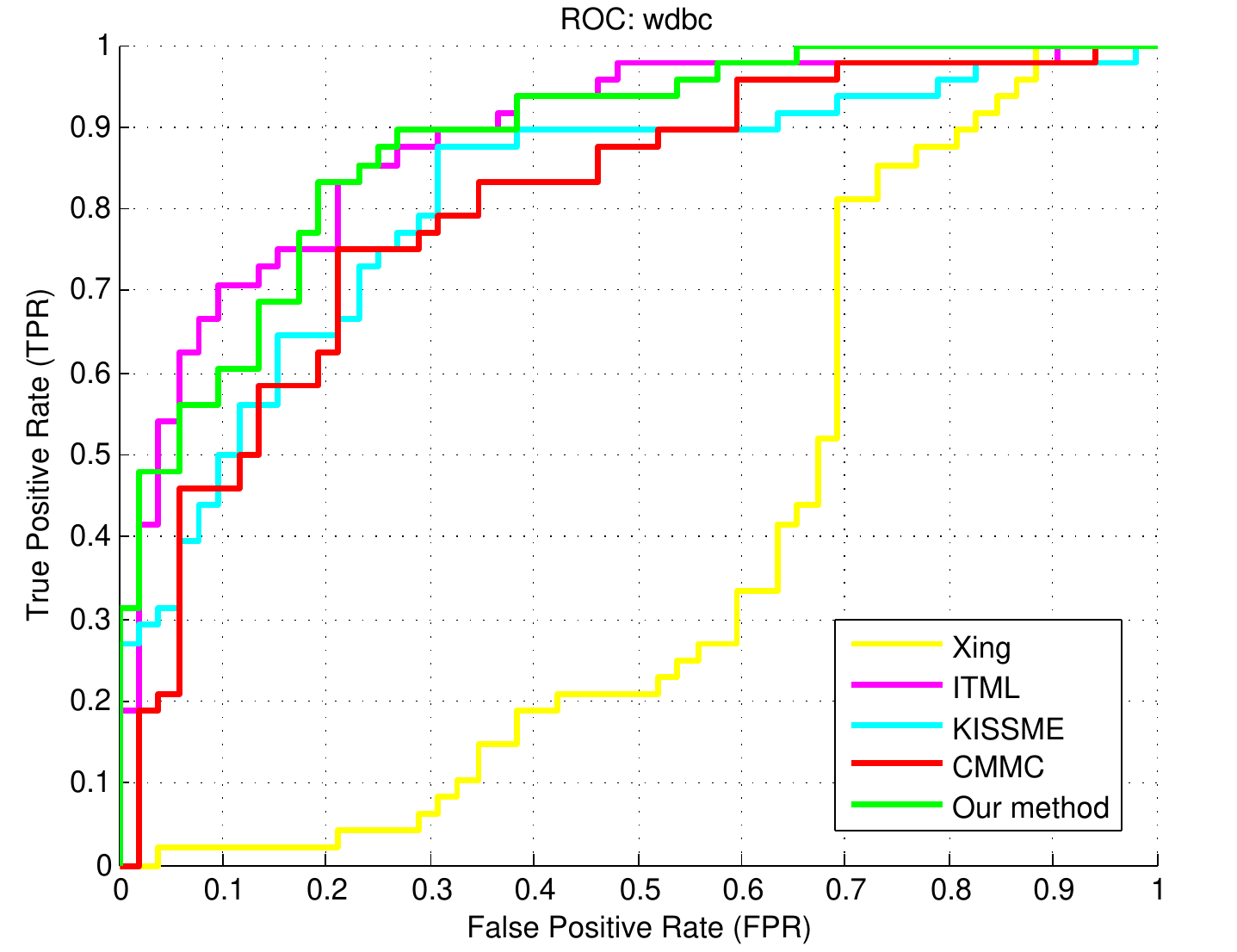}&
    \includegraphics[trim = 35mm 83mm 35mm 84mm, clip, width=6.5cm]{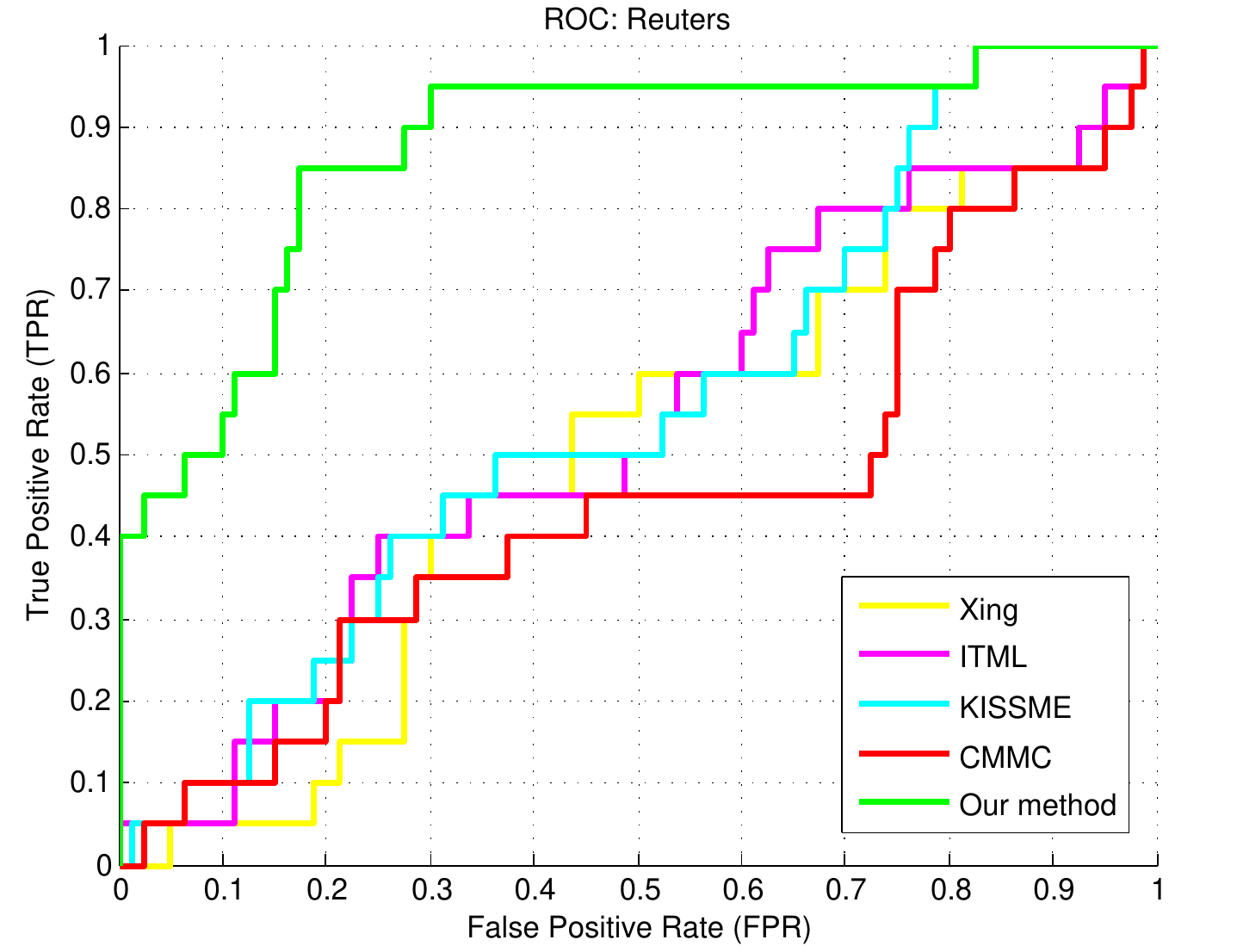}\\
  \end{tabular}
\caption{The pairwise classification results on the five UCI data sets and the Reuters data set (the right bottom).}
\label{fig:uci}
\end{figure}

{\bf UCI data sets}: In the experiment, we selected the five widely used data sets from the UCI machine learning repository\footnote{\url{https://archive.ics.uci.edu/ml/datasets.html}}, which has different dimension and categories, shown in Table \ref{tab:dataset}. As for the number of hidden units in our model, we set the number of hidden nodes to be 100 on the sonar data set and 64 on the other UCI data sets. For the each data set, we randomly sampled 200 pairwise constraints, of which 100 pairs were used for training and the rest to test the pairwise classification performance. While for clustering performance, we test the model on all the data elements. We compared our method to the state of the art methods, and clustering results are shown in Table. (\ref{tab:tab1}). It demonstrates that our method outperforms other methods on almost all the data sets, especially for the data with the larger number of classes. We also show the performance of our method on the pairwise classification task in Fig. \ref{fig:uci}. Except on the Wdbc and glass data sets, our method yields completive and even better results than other methods.

{\bf Reuters data set}:  We used the Reuters21578\footnote{\url{http://www.cad.zju.edu.cn/home/dengcai/Data/TextData.html}}, which has the total 8293 documents with 18933 dimensional features for each document, belonging to 65 categories. Because the Reuters data set has high dimension, we first projected it into 400 dimensions with PCA. Then we set the number of hidden nodes to be 100 in our model. The clustering performance is shown in Table. (\ref{tab:tab1}). It demonstrates that our method is significantly better than other methods. Again, our method yields remarkably better pairwise classification result, shown in the right bottom of Fig. \ref{fig:uci}.

  \begin{figure}[htp]
  \centering
  \begin{tabular}{cc}
    \includegraphics[trim = 35mm 84mm 40mm 88mm, clip, width=6.5cm]{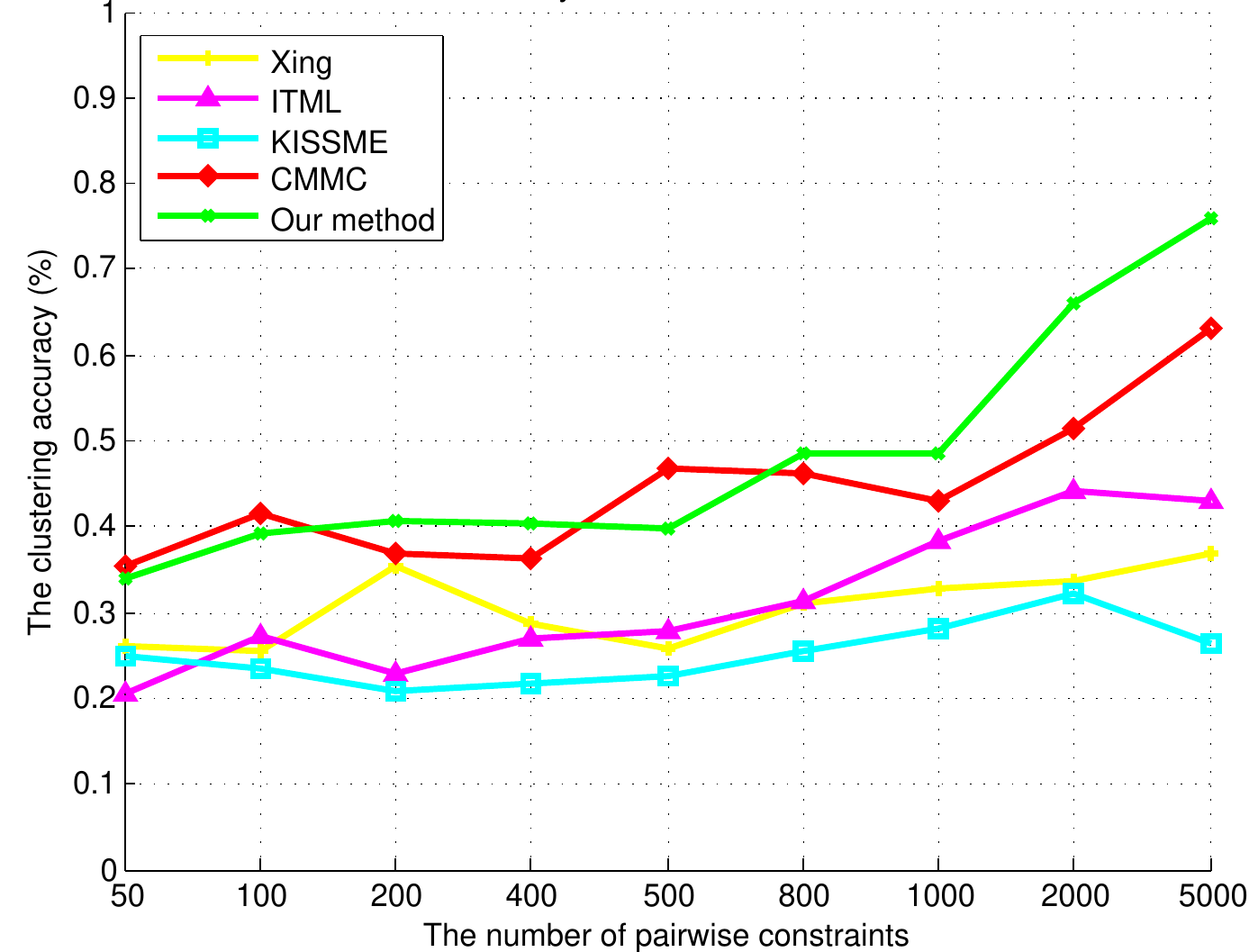}&
    \includegraphics[trim = 35mm 84mm 40mm 88mm, clip, width=6.5cm]{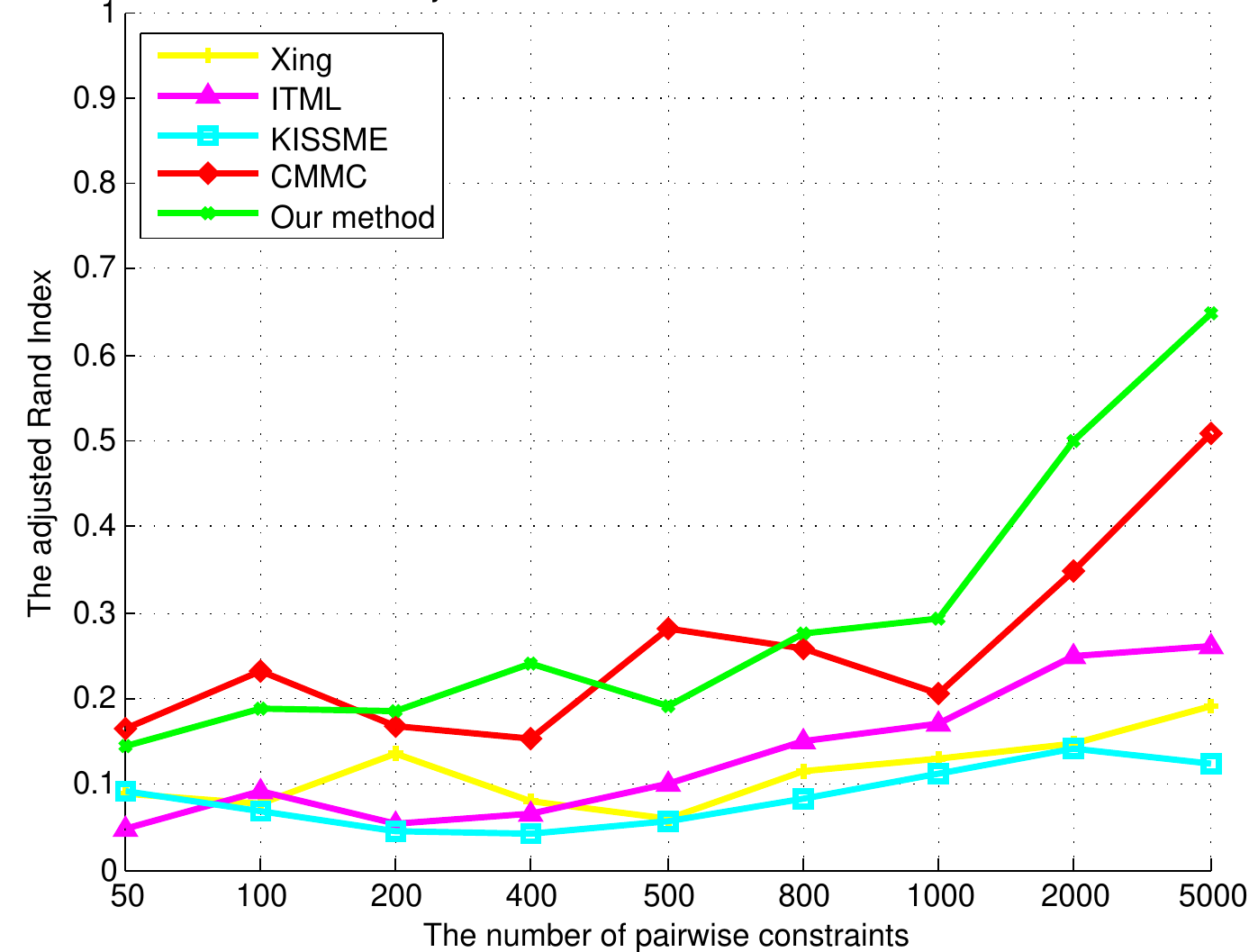}\\
  \end{tabular}
\caption{The clustering performance comparison on the MNIST digits by varying the number of training pairs, evaluated with accuracy and rand index respectively. It demonstrates that our method is significantly better than other methods for clustering analysis.}
\label{fig:mnist}
\end{figure}


 
{\bf MNIST dataset}: The MNIST digits\footnote{\url{http://yann.lecun.com/exdb/mnist/}} consists of $28\times28$-size images of handwriting digits from $0$ through $9$ with a training set of 60,000 examples and a test set of 10,000 examples, and has been widely used to test character recognition methods. In the experiment, we randomly sampled a subset with 5000 images from the training sets to test our method and other baselines. In the experiment, we use a three-layer deep structure for MNIST digits, with hidden nodes [400 200 100] respectively on each layer. We tested how the clustering performance changes when the number of pairwise constraints varies. The experimental comparisons between our method and other baselines are shown in Fig. \ref{fig:mnist}. It demonstrates that the clustering accuracy is increasing with more pairwise constraints. And it also shows our method is better than other baselines in most cases when varying the number of training pairs. 

\begin{figure}[htp]
  \centering
  \begin{tabular}{cc}
    \includegraphics[trim = 35mm 84mm 40mm 88mm, clip, width=6.5cm]{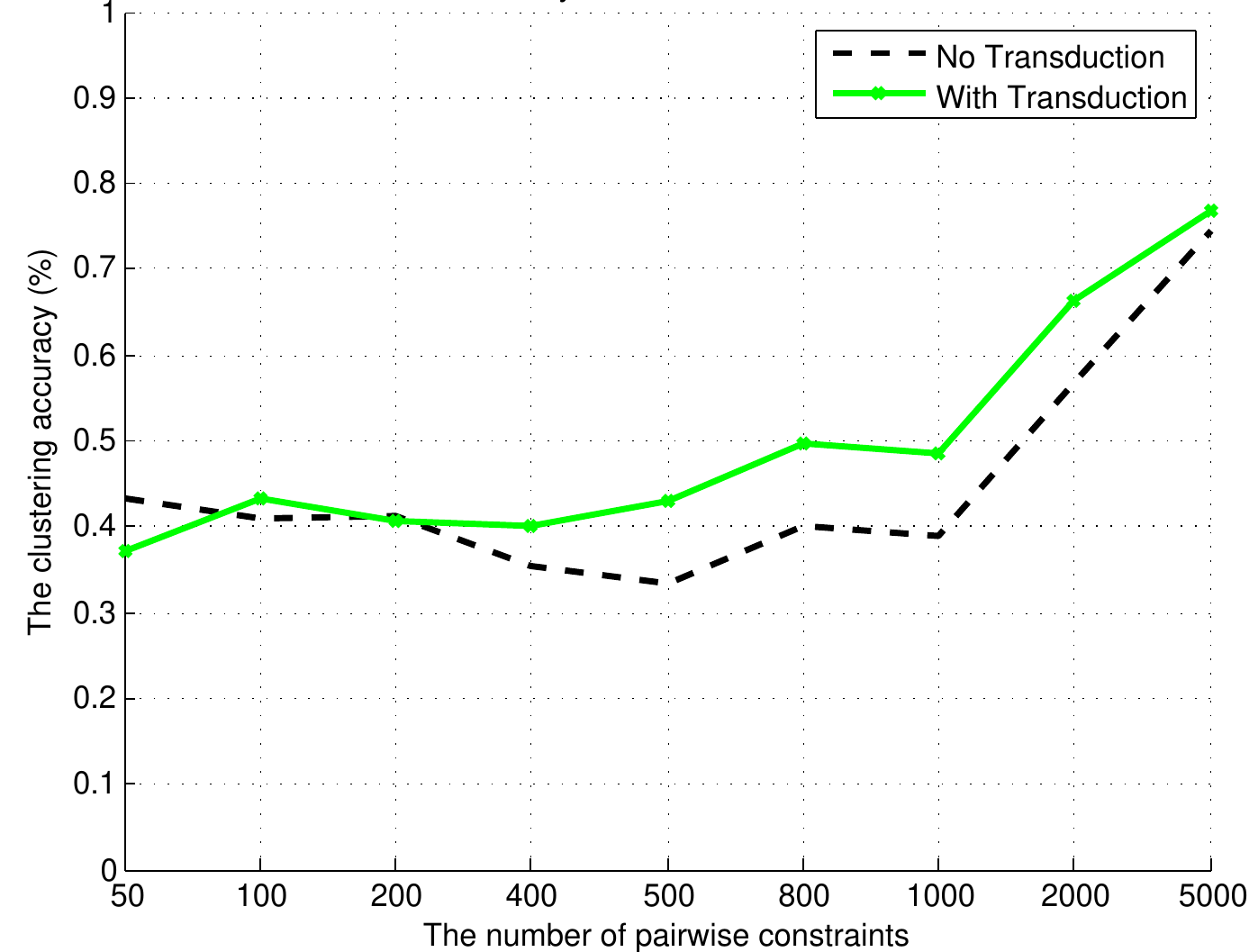}&
    \includegraphics[trim = 35mm 84mm 40mm 88mm, clip, width=6.5cm]{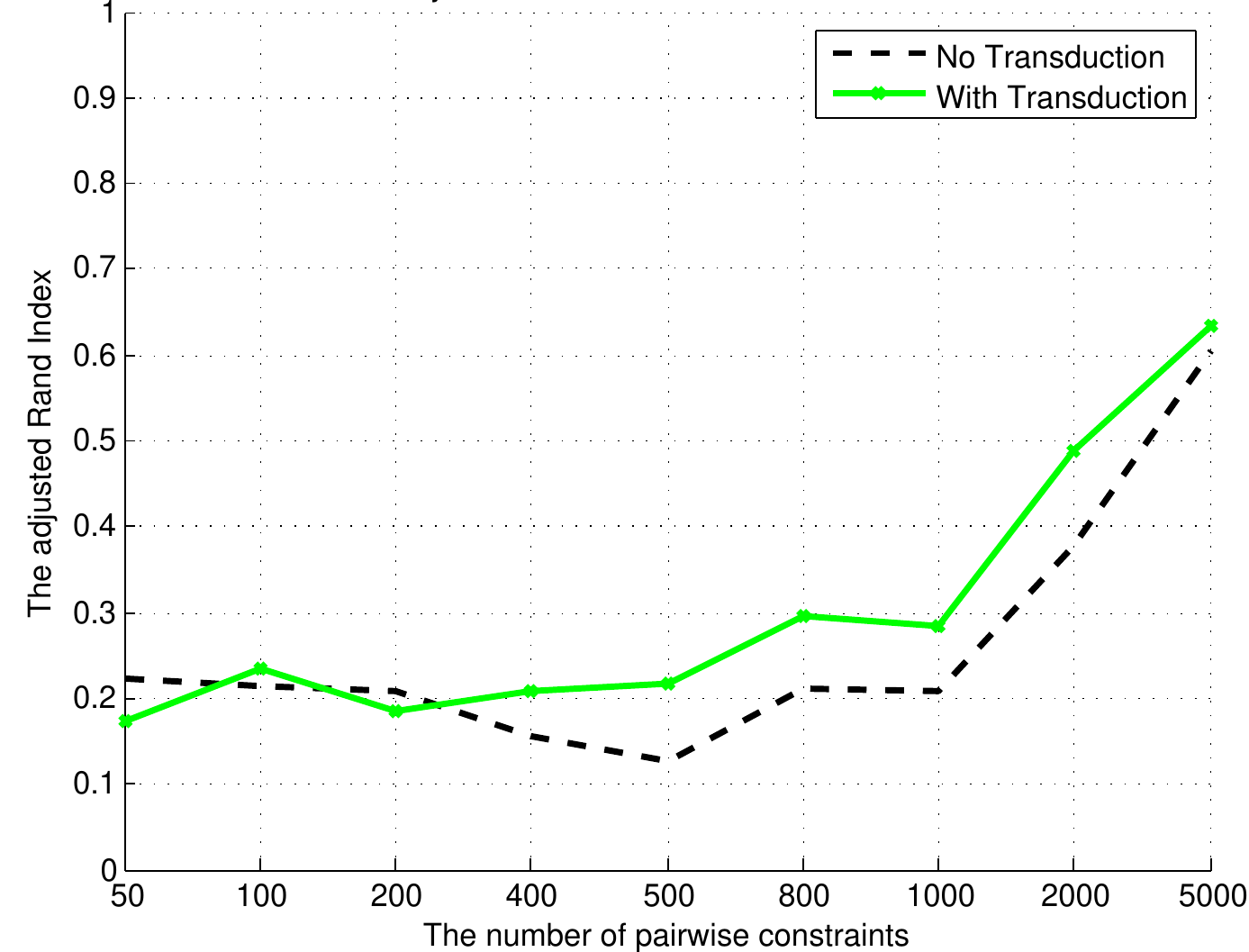}\\
  \end{tabular}
\caption{The comparison between with and without transductive principles for our method. (a) and (b) show the results (evaluated with accuracy and rand index respectively) on the MNIST data set.}
\label{fig:mnist_tran}
\end{figure}

To evaluate whether the transductive constraint in our model in Eq. \ref{eq:obj} is helpful or not for clustering, we set $\beta = 0$ to get rid of the transductive condition in Eq. \ref{eq:cond3}, and the experimental results are shown in Fig. \ref{fig:mnist_tran}. We argue that the result in Fig. \ref{fig:mnist_tran} is consistent with common sense. The smaller the number of pairwise constraints, the higher uncertainty when we do inference. Thus, transductive learning has no advantage when the number of constraints is small. But it performs better with more constraints in Fig. \ref{fig:mnist_tran}. When more and more pairwise constraints are available, there's no need to incorporate transductive principles in the model. To sum up, it demonstrates that the transductive constraint in our model is remarkably helpful for the semi-supervised clustering analysis. 

{\bf COIL data set}: We test our method on both COIL-20 and COIL-100 image data sets. The COIL-20 data set\footnote{\url{http://www.cs.columbia.edu/CAVE/software/softlib/coil-20.php}} has total 1440 images, with size $128\times128$. It is divided into 20 classes of objects, with 72 images for each object. In our experiments, we used the processed version, which contains images for all of the objects in which the background has been discarded, and furthermore we resized all the images into $32\times32$ for space and time concern. The COIL-100 data set consists of 7200 images, partitioned into 100 classes. Similarly, we also resized the images into $32\times32$ before learning clustering model. 

\begin{figure*}[th!]
  \centering
  \begin{tabular}{cccc}
    \includegraphics[trim = 37mm 80mm 40mm 80mm, clip, width=4.0cm]{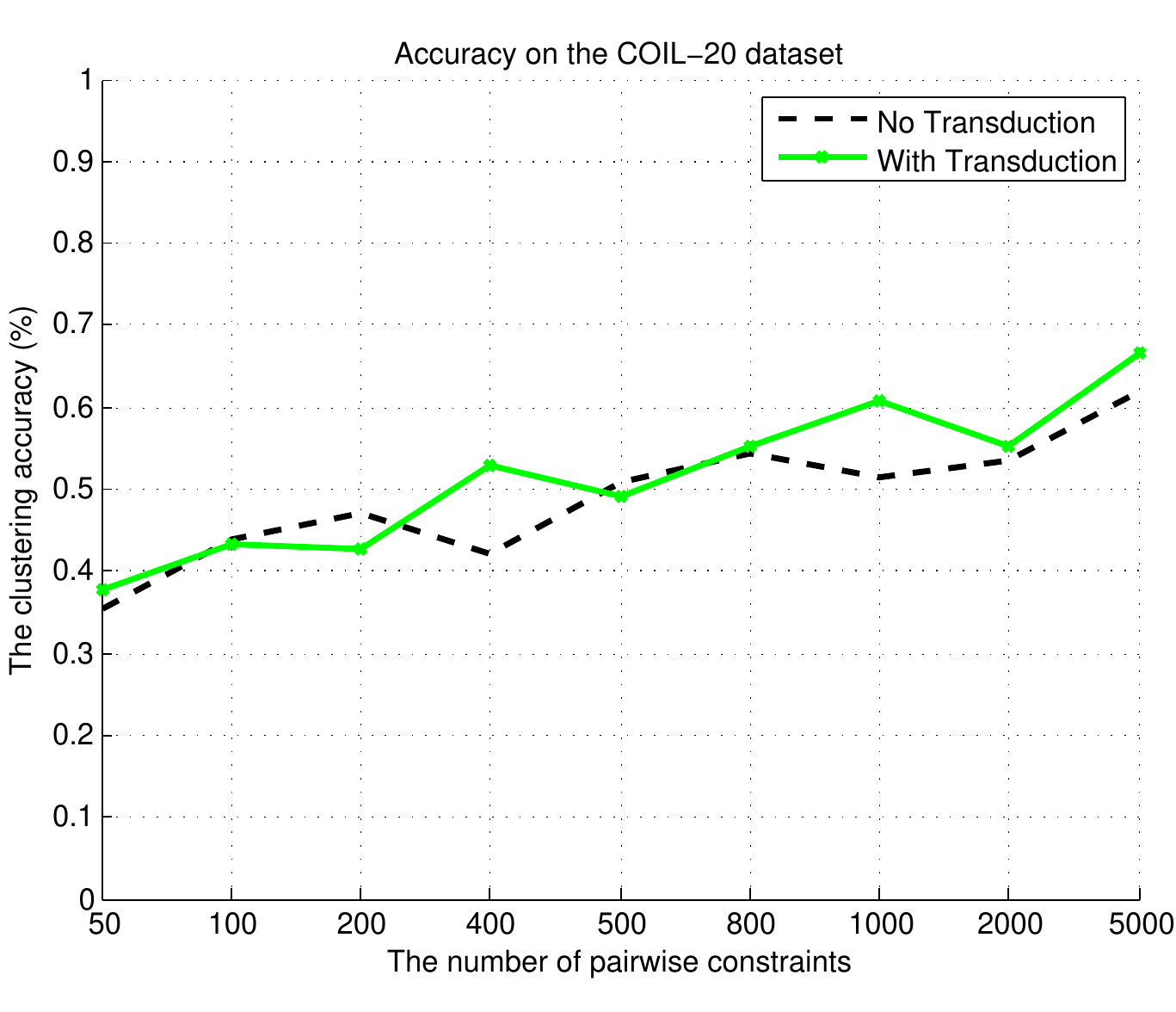} &
    \includegraphics[trim = 37mm 80mm 40mm 80mm, clip, width=4.0cm]{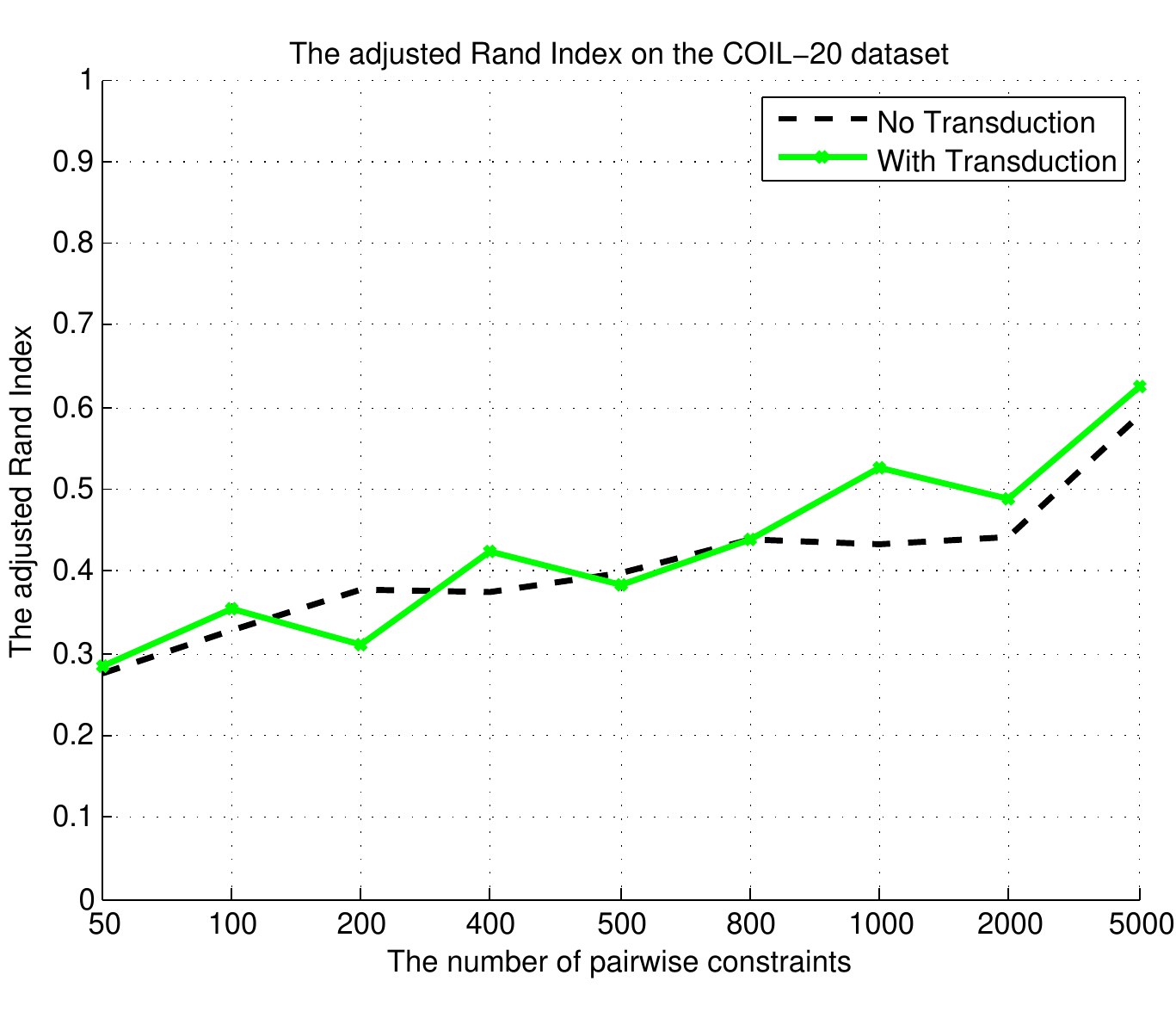}& 
        \includegraphics[trim = 36mm 80mm 40mm 80mm, clip, width=4cm]{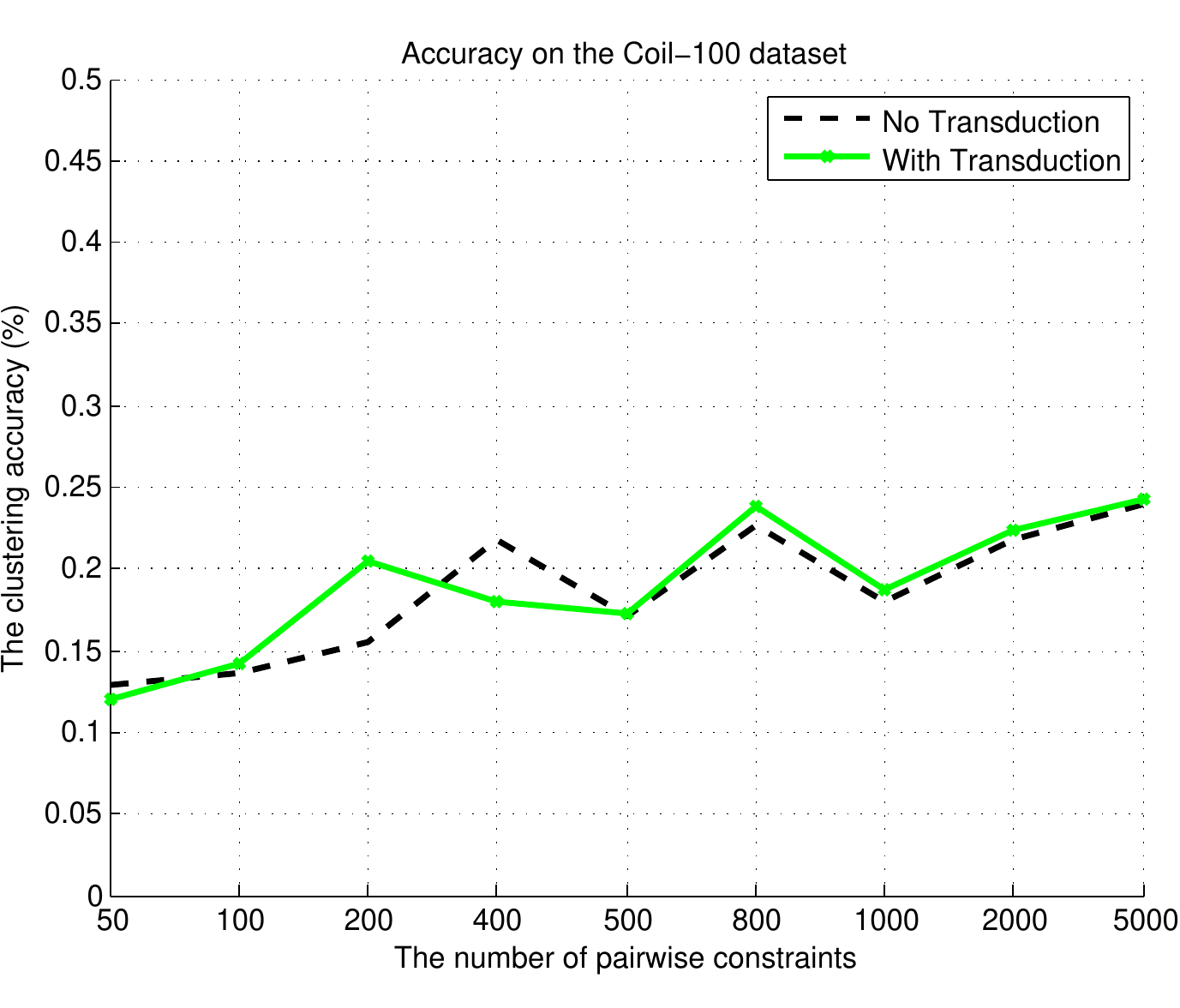}&
    \includegraphics[trim = 36mm 80mm 40mm 80mm, clip, width=4cm]{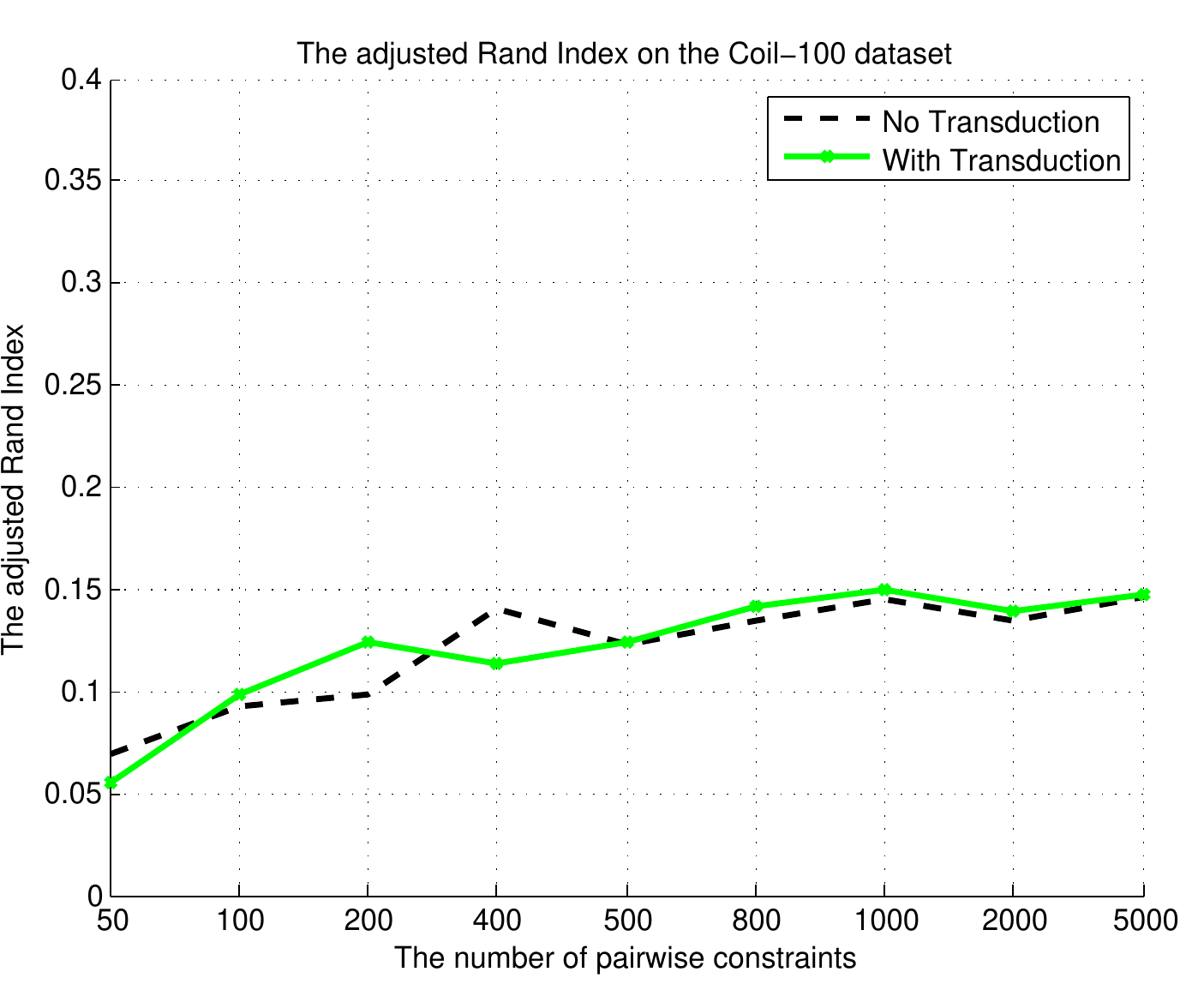}\\
    (a) & (b) & (c) & (d)
  \end{tabular}
\caption{The comparison between with and without transductive principles for our method. (a) and (b) show the results (evaluated with accuracy and rand index respectively) on the COIL-20 data set; (c) and (d) are the results on the COIL-100 dataset, with accuracy and rand index respectively. For the 20 classes, it shows that the transductive learning is helpful when the number of training pairs is small. However, for the 100 classes, the transductive learning cannot improve the performance. It demonstrates that it is better to leverage transductive principles when the number of classes is relative smaller.}
\label{fig:coil100_tran}
\end{figure*}

\begin{figure}[htp]
  \centering
  \begin{tabular}{cc}
    \includegraphics[trim = 20mm 70mm 20mm 68mm, clip, width=6.5cm]{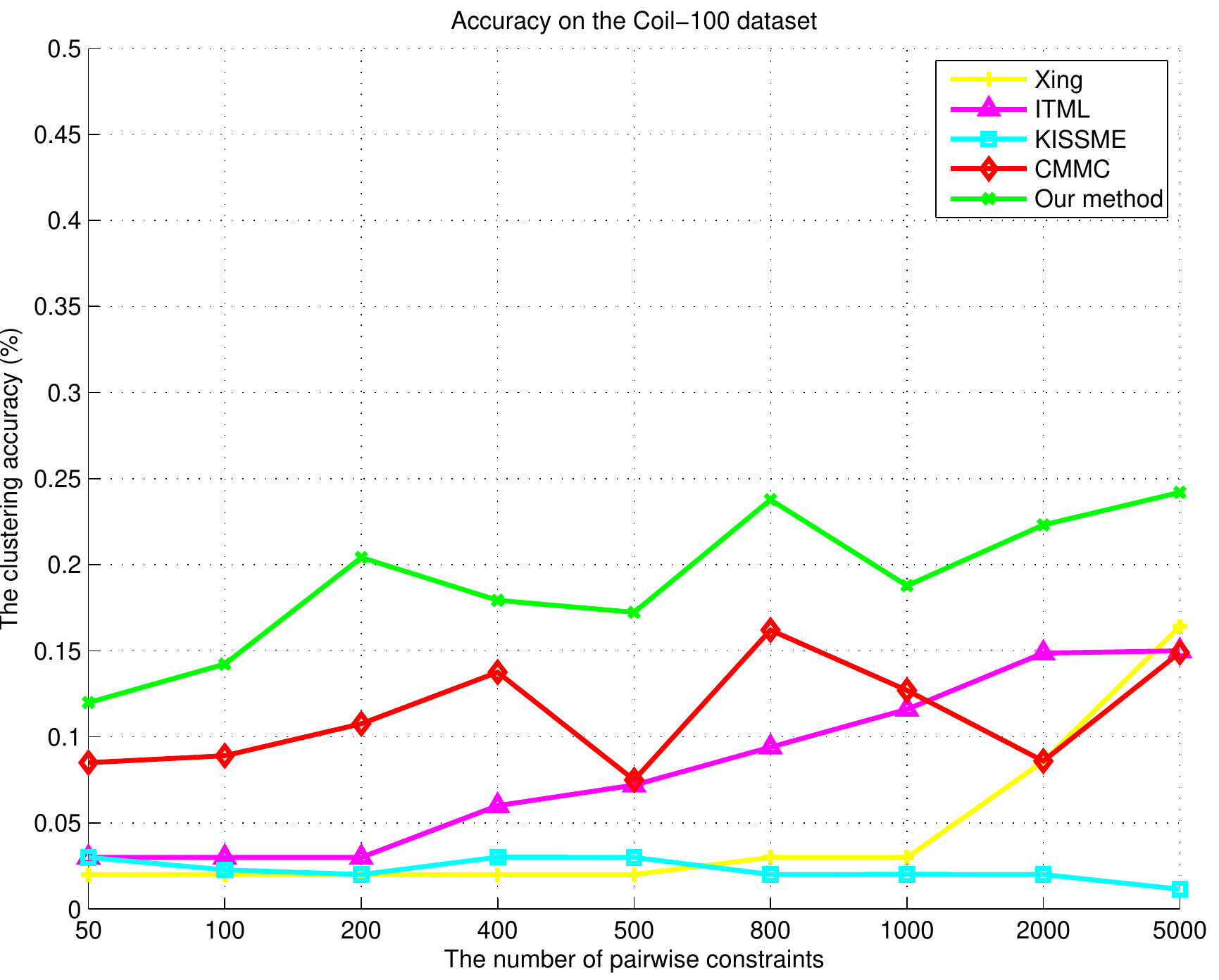}&
    \includegraphics[trim = 20mm 70mm 20mm 68mm, clip, width=6.5cm]{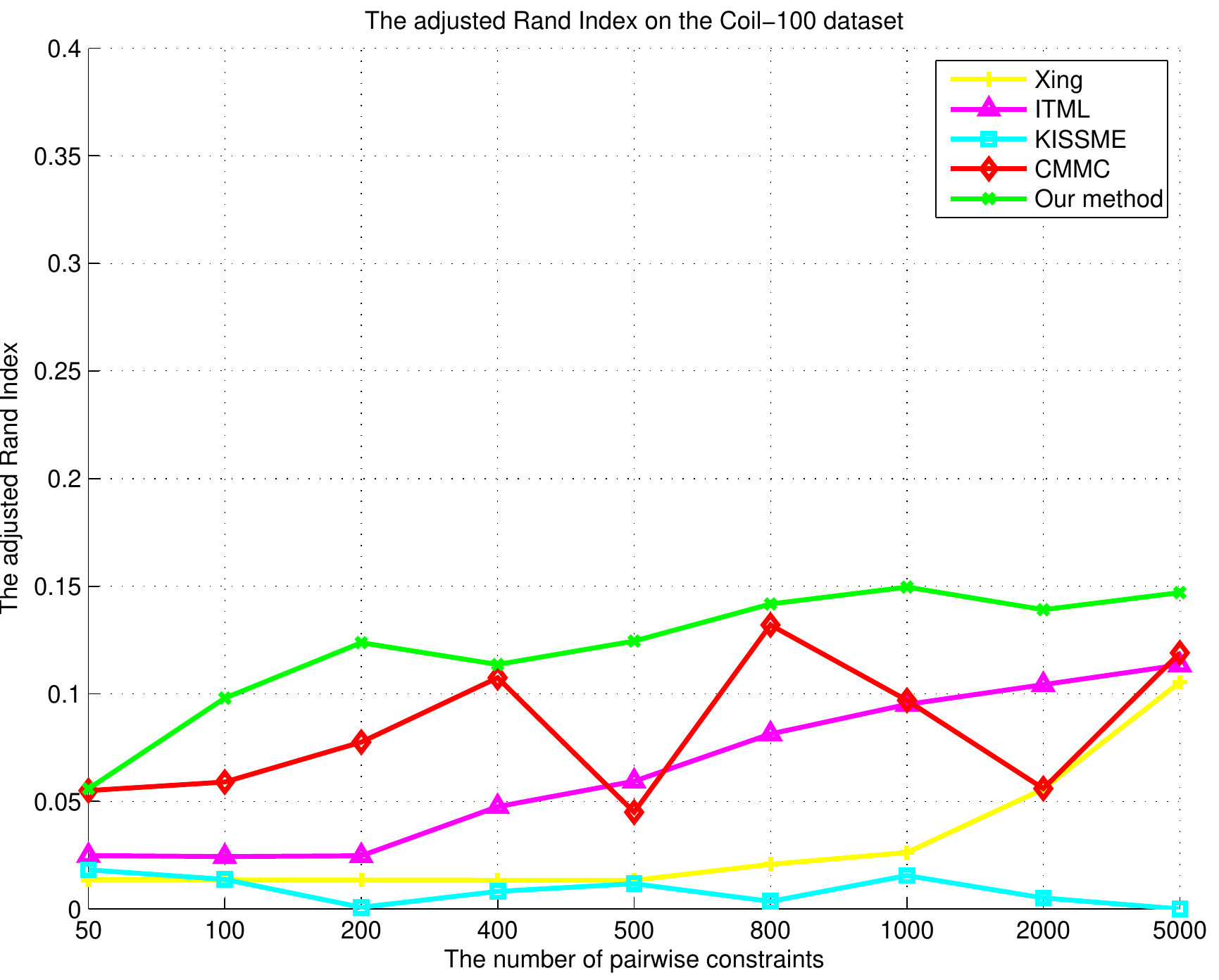}\\
  \end{tabular}
\caption{The clustering comparison on the COIL-100 data set by varying the number of training pairs. It shows that our method outperforms other methods significantly.}
\label{fig:coil100}
\end{figure}

The clustering performance is shown in Fig. \ref{fig:coil100}, and it demonstrates that our method is better than other methods with both stability and clustering accuracy. Where's the performance gain from in Fig. \ref{fig:coil100}? deep learning or transductive learning? To answer this question, we evaluated whether transductive training is helpful when the number of pairwise constraints is limited. In Figs. \ref{fig:coil100_tran}(a) and (b), it shows the results on 20 classes, and demonstrates that transductive learning is helpful when the number of training pairs is in range [400 2000]. But with more and more training constraints, transductive learning cannot improve the performance too much. In Figs. \ref{fig:coil100_tran}(c) and (d), it shows the results on COIL with 100 classes, and indicates that transductive learning cannot improve the performance much when the number of classes is large. We think the reason that transductive learning cannot perform well in Figs. \ref{fig:coil100_tran}(c) and (d) is that it cannot infer label well with large margin on the dataset with a larger number of clusters. We argue when we have more data and more clusters, it is harder to partition the data well, and more difficult to find a better hyperplane to separate one cluster well from the others with large margin. In other words, it is harder to satisfy the condition in Eq. \ref{eq:cond3}. Compared to the COIL dataset, transductive learning yields a larger gain on the MNIST data set in Fig. \ref{fig:mnist_tran}. Thus, transductive learning is helpful when the number of classes is small and the data is well distributed (compact within the same cluster, and separated between different clusters).

Thus, the most performance gain on the COIL-100 data set in Fig. \ref{fig:coil100} is from deep learning, according to the above analysis. 


%

%
%
\section{Conclusions}
In this paper, we propose a deep transductive semi-supervised maximum margin clustering approach. One the one hand, we leverage deep learning to learn non-linear representations, which can be used as the input to the semi-supervised clustering model. On the other hand, we incorporate the non-label instances into our semi-supervised clustering framework. Thus, our model unifies transductive learning, deep learning, maximum margin and semi-supervised clustering in one framework. Compared to conventional methods, our approach can learn non-linear mappings as well as leveraging transductive information to improve clustering performance. We pretrain the deep structure with stacked restricted Boltzmann machines layer by layer greedily for feature representations and optimize our objective function with gradient decent. We demonstrate the advantages of our model over the state of the art in the experiments.

\bibliographystyle{ieee}
\bibliography{rbmbib}

\end{document}